\documentclass[letterpaper, 10 pt, conference]{ieeeconf}  % Comment this line out if you need a4paper

\IEEEoverridecommandlockouts                              % This command is only needed if 
\usepackage{amsmath}
\usepackage{amssymb}
\usepackage{graphicx}
\usepackage{xcolor}
\usepackage{subcaption}
\usepackage{dashrule}
\usepackage{booktabs}
\usepackage{array}
\usepackage{tikz}
\usepackage{multirow}
\title{\LARGE \bf
Dense Force Estimation with an Event-based Optical Tactile Sensor
}

\author{Agis Politis$^{1}$, René Zurbrügg$^{2}$, Valentina Cavinato$^{1}$% <-this % stops a space
\thanks{$^{1}$Agis Politis and $^{1}$Valentina Cavinato are with Sony Advanced Visual Sensing, Zurich, Switzerland,
        {\tt\small Agis.Politis@sony.com}.
        René Zurbrügg is with the Computer Vision and Geometry Group at ETH Zürich.}%
}

\begin{document}

\maketitle
\thispagestyle{empty}
\pagestyle{empty}

%%%%%%%%%%%%%%%%%%%%%%%%%%%%%%%%%%%%%%%%%%%%%%%%%%%%%%%%%%%%%%%%%%%%%%%%%%%%%%%%
\begin{abstract}
Humans rely on spatially dense, geometry and force-aware tactile feedback at high temporal resolution for dexterous manipulation. While vision-based tactile sensors enable dense force estimation, they are limited by camera frame rates, motion blur, and data bandwidth. Event-based optical tactile sensors offer an attractive alternative with microsecond temporal resolution and low motion blur, but existing methods are restricted to predicting only net forces. We introduce the first framework for dense 3D force field reconstruction using event-based optical tactile sensors. Our approach estimates 3D surface displacements from event data and maps them to forces via the inverse Finite Elements Method (iFEM). Shear displacements are recovered through the proposed event-based marker tracking algorithm, while normal displacements are predicted by a convolutional neural network trained on a collected dataset of synchronized force-displacement-event data. Experiments demonstrate accurate reconstruction of physically grounded forces, achieving a mean absolute error of (0.14 N, 0.10 N, 0.93 N) over force ranges up to (4 N, 4 N, 20 N), while operating at an average of 100 Hz. This work constitutes a first step toward enabling dense force feedback for high-frequency control in robotic grasping and dexterous manipulation.

\end{abstract}

%%%%%%%%%%%%%%%%%%%%%%%%%%%%%%%%%%%%%%%%%%%%%%%%%%%%%%%%%%%%%%%%%%%%%%%%%%%%%%%%
\section{INTRODUCTION}

Touch is fundamental to how humans interact with the world. By sensing rich contact properties at high spatio-temporal resolutions, such as the magnitude and distribution of contact forces, humans achieve dexterous manipulation \cite{johansson2009coding}. In robotics, replicating such capabilities is essential for safe, adaptive, and precise physical interaction, complementing sensing modalities such as vision. Recent work highlights that dense tactile forces provide compact and informative feedback for robotic manipulation tasks \cite{luu2025manifeel}.

Many tactile sensors have been developed to equip robots with the sense of touch. Capacitive and magnetic sensors offer high-frequency feedback \cite{bhirangi2021reskin}, but typically lack spatial resolution. Vision-based alternatives can infer dense force information by observing the deformation of a clear elastomer \cite{johnson2009retrographic}, but are limited by camera frame rates, motion blur, and bandwidth. Event cameras provide a promising alternative \cite{kumagai2019event, funk2024evetac, yin2025gelevent}, offering microsecond temporal resolution, low latency, and minimal blur \cite{gallego2020event}, however, existing event-based tactile methods are limited to estimating net forces.

\begin{figure}[t]
\centering
\setlength{\tabcolsep}{0.8pt}

% regular panel height
\newlength{\imgH}
\setlength{\imgH}{0.25\linewidth}

% explicit vertical gap between the two rows
\newlength{\rowgap}
\setlength{\rowgap}{1pt}

% target height of the tall left panel = two image heights + one row gap
\newlength{\imgVH}
\setlength{\imgVH}{\dimexpr 2\imgH + \rowgap\relax}

% independently adjustable size of the tall left panel
\newlength{\leftImgW}
\setlength{\leftImgW}{0.25\linewidth}   % adjust width here
\newlength{\leftImgH}
\setlength{\leftImgH}{\dimexpr\imgVH+4.15pt\relax}           % adjust height here, e.g. \dimexpr\imgVH+2pt\relax

\newlength{\colgap}
\setlength{\colgap}{2.0pt} % try 2–4pt for visibility in PPT

% helper: standard image with label
\newcommand{\imglabh}[3]{%
\begin{tikzpicture}
\node[inner sep=0pt, outer sep=0pt, anchor=south west] (img)
{\includegraphics[height=#3,keepaspectratio]{#1}};
\node[anchor=north west, xshift=1pt, yshift=-1pt]
at (img.north west) {\color{white}\scriptsize\textbf{(#2)}};
\end{tikzpicture}%
}

% helper: fully adjustable image with label
\newcommand{\imglabstretch}[4]{%
\begin{tikzpicture}
\node[inner sep=0pt, outer sep=0pt, anchor=south west] (img)
{\includegraphics[width=#3,height=#4]{#1}};
\node[anchor=north west, xshift=1pt, yshift=-1pt]
at (img.north west) {\color{white}\scriptsize\textbf{(#2)}};
\end{tikzpicture}%
}

\begin{tabular}{@{}c@{\hspace{\colgap}}ccc@{}}
% tall left panel
\raisebox{-66.15pt}[0pt][0pt]{%
\imglabstretch{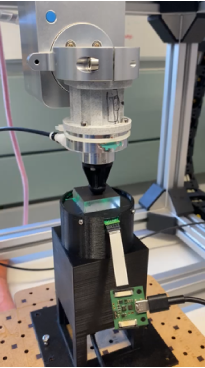}{a}{\leftImgW}{\leftImgH}} &
\imglabh{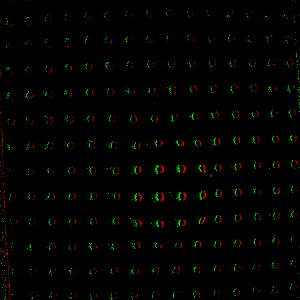}{b}{\imgH} &
\imglabh{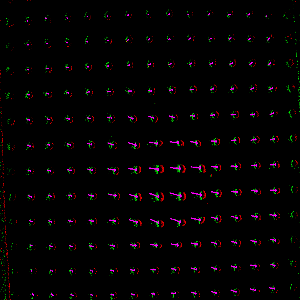}{c}{\imgH} &
\imglabh{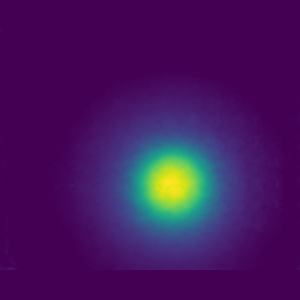}{d}{\imgH} \\[\rowgap]

&
\imglabh{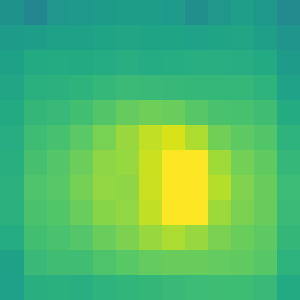}{e}{\imgH} &
\imglabh{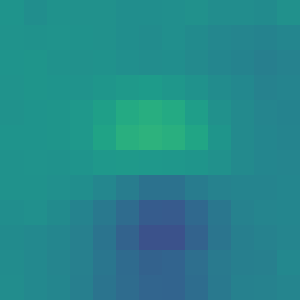}{f}{\imgH} &
\imglabh{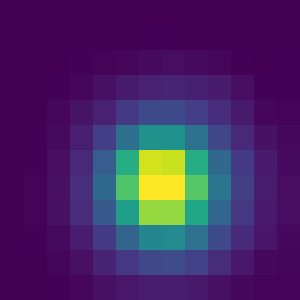}{g}{\imgH} \\
\end{tabular}

\caption{(a) Tactile sensor and probing setup, (b) Surface of active events, (c) Marker tracks, (d) Predicted normal deformation, (e) X-axis shear forces, (f) Y-axis shear forces, (g) Normal forces. Forces shown in marker resolution.}
\label{fig:six_panel}
\vspace{-1.3em}
\end{figure}
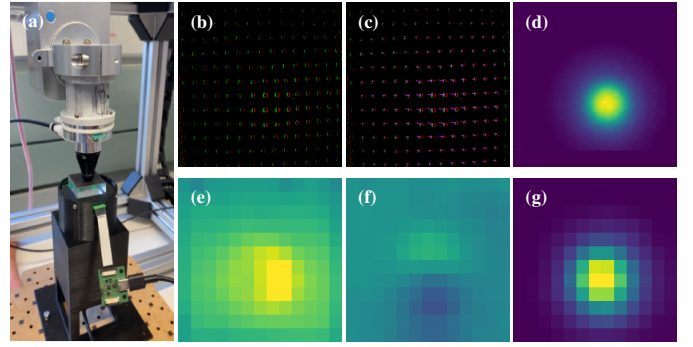

Dense force reconstruction has been widely studied in RGB tactile sensing. Model-based methods leverage iFEM \cite{ma2019dense, zhao2024ifem2} or force decomposition-based approaches \cite{zhang2022deltact}, while learning-based methods directly regress forces from tactile images \cite{sferrazza2019ground}. However, these approaches rely on dense colored tactile imprints that directly encode surface deformation. In contrast, event cameras generate sparse and asynchronous measurements driven primarily by marker motion, making normal deformation not directly observable, and thus the estimation of dense geometry-aware signals such as depth and forces challenging.

We address this challenge by presenting, to the best of our knowledge, the first pipeline for dense 3D force reconstruction using an event-based optical tactile sensor. Our approach computes 3D marker displacements by combining event-based marker tracking for shear estimation, with a neural network that predicts dense normal deformation. The resulting displacement field is converted into physically grounded forces using iFEM~\cite{ma2019dense}. On a dataset with four indenter shapes and forces up to $(4\,\mathrm{N}, 4\,\mathrm{N}, 20\,\mathrm{N})$, we achieve mean absolute errors of $(0.14\,\mathrm{N}, 0.10\,\mathrm{N}, 0.93\,\mathrm{N})$ for shear x, y and normal forces, while operating at 100\,Hz on average. Our contributions are summarized as follows:

\begin{itemize}
    \item We introduce the first dense 3D force reconstruction method for event-based optical tactile sensing, achieving high-frequency operation at 100\,Hz.
    \item We propose a learning-based method to infer dense normal deformation from sparse asynchronous event data, addressing a challenge of event-based tactile sensors.
    \item We collect and release a dataset of synchronized force-displacement-event recordings with diverse contacts.
\end{itemize}

\begin{figure*}[]
    \centering
    \includegraphics[width=0.9\textwidth]{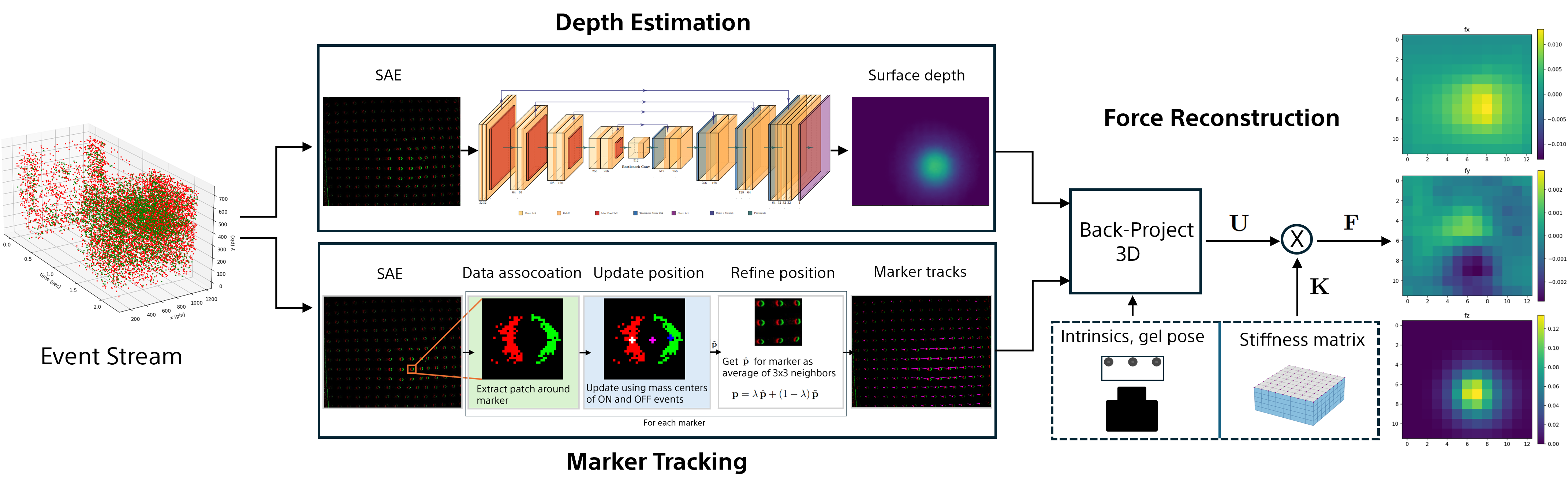}
    \caption[Dense 3D force reconstruction pipeline.]{Dense 3D force reconstruction method: (a) Marker tracking for shear displacements, (b) Depth estimation for normal displacements, (c) Mapping marker displacements to forces, through iFEM. Dashed lines are components computed offline.}
    \label{fig:main_pipeline}
    \vspace{-1.8em}
\end{figure*}

\section{HARDWARE AND SENSOR CALIBRATION}

Our prototype follows a similar design to \cite{funk2024evetac}. It features an event camera  (Prophesee EVK4 with Sony IMX636 event vision sensor) equipped with an adjustable C-mount lens, an LED ring with a PCB for uniform illumination, and a transparent elastomer covered by black membrane with embedded markers in a \(12\times13\) regular grid with 2\,mm pitch. %Under external contact, marker motion causes brightness changes, which are captured as events by the camera and used as sensing features.

Our method requires mapping tracked 2D motions to 3D displacements, thus, the projective transformation between gel and camera must be known. Camera intrinsics are obtained using the method of~\cite{muglikar2021calibrate} with the E2VID~\cite{rebecq2019high} network, while extrinsics are estimated from accumulated event frames generated during LED blinking, yielding a mean reprojection error of $0.64 \pm 0.24$ pixels.

\vspace{-0.2em}

\section{METHOD}
We tackle the problem of estimating a 3D force field acting on the surface of a gel, using measurements from an event-based optical tactile sensor. Under contact, we aim to recover the 3D forces $\mathbf{f}_j \in \mathbb{R}^3$ acting on surface markers $j=1,\dots,M$, forming the field $\mathbf{F}=\{\mathbf{f}_j\}_{j=1}^M$. Assuming a linear elastic gel with known material parameters, we recover the 3D marker displacements $\mathbf{U}=\{\mathbf{u}_j\}_{j=1}^M$ from the event stream and map them to forces through iFEM (Figure \ref{fig:main_pipeline}).

\vspace{-0.2em}
\subsection{MARKER TRACKING}
\label{sec:marker_tracking}

We develop a lightweight algorithm that operates on short event batches and maintains marker tracks over time. During contact, most events are generated by moving markers, and motion between consecutive updates is small if the time window is small. These properties allow for reliable local data association: events in a small marker's neighborhood are likely caused by that marker and form a ring-like pattern (Figure \ref{fig:main_pipeline}). Tracking is initialized by projecting 3D marker locations to the image plane using the camera matrix.

At each step $t$, events are accumulated into polarity-separated Surface of Active Events (SAE) \cite{gehrig2019end}. For each marker, we extract a square patch centered at its previous position $\mathbf{p}^t$ and estimate motion by computing a centroid of events within this region, producing an updated position $\tilde{\mathbf{p}}^{\,t}$. Under large deformations, neighboring markers may approach, causing patch overlap and incorrect updates. To prevent this, we restrict patches to non-overlapping areas, defined by the midpoints between adjacent markers. Although this update is usually effective, tracking may still degrade when spurious events arise (e.g. from object edges) or when neighboring markers move very close. To improve robustness, we introduce a grid-aware refinement step. Each updated position is combined with a prior position $\hat{\mathbf{p}}^{\,t}$ obtained from the average location of its immediate neighbors,
\begin{equation}
\mathbf{p}^{\,t+1} = \lambda\,\hat{\mathbf{p}}^{\,t} + (1-\lambda)\,\tilde{\mathbf{p}}^{\,t}
\end{equation}
where $\lambda \in [0,1]$ controls the prior's strength. Tracking is skipped when a limited number of events are present in a marker's patch to prevent artificial drift under no motion.

\subsection{ESTIMATING NORMAL DISPLACEMENT}
\label{sec:depth_estimation}

To recover the 3D deformation field, we must also estimate the normal displacement of the gel surface, or its depth. We investigate whether this is feasible using the sparse and asynchronous signals produced by an event-based tactile sensor. To this end, we develop a neural network that predicts dense per-pixel normal deformation from short event windows, trained on the dataset described in Sec.~\ref{sec:data_collection}.

We use a U-Net~\cite{ronneberger2015u} (7.9M parameters) to regress depth maps from recent events. The input is a polarity-separated SAE, normalized to \([0,1]\), resized and cropped to the gel region forming a \(2\times90\times112\) tensor. The output is a single-channel depth map of matching resolution. This representation is motivated by~\cite{zhang2022deltact}, who showed that global marker patterns encode sufficient information to infer coarse normal deformations. We observe that even localized contacts usually induce motion across all markers, generating events that implicitly encode the global deformation of the gel surface.

Unlike conventional supervised learning with image-label pairs, our data consist of quasi-continuous event streams synchronized with high-frequency ground-truth. For training, we sample 10~ms event windows, see (Sec.~\ref{sec:depth_eval}), and the corresponding depth maps and contact masks. Specifically, a recording and a temporal index within the recording are first sampled at random, and all events within the subsequent 10~ms are accumulated into an SAE. The sampled ground-truth is the one nearest in time to the last event of the batch, ensuring temporal alignment with a maximum offset of 5~ms. Based on the maximum indentation speed, this offset can result in a maximum error of 0.05 mm. This sampling method leverages the temporal extent of the data, enhancing data diversity and promoting generalization. We adopt a $\mathcal{L}_1$ loss between the predicted and ground-truth depth. To make the loss invariant to the indenter’s contact area and avoid bias toward predicting zero deformation in small contact cases, we compute it separately over contact and non-contact regions and combine with a weighting factor:

\vspace{-1.5em}

\begin{subequations}
\label{eq:loss_block}
\begin{align}
\mathcal{L}_{\text{contact}} &= \tfrac{1}{|\mathcal{C}|}\!\sum_{\mathbf{x}\in\mathcal{C}} \! \big| D_{\text{pred}}(\mathbf{x}) - D_{\text{gt}}(\mathbf{x}) \big| \\
\mathcal{L}_{\text{noncontact}} &= \tfrac{1}{|\bar{\mathcal{C}}|}\!\sum_{\mathbf{x}\in\bar{\mathcal{C}}} \! \big| D_{\text{pred}}(\mathbf{x}) - D_{\text{gt}}(\mathbf{x}) \big| \\
\mathcal{L} &= \lambda_c\,\mathcal{L}_{\text{contact}} + (1-\lambda_c)\,\mathcal{L}_{\text{noncontact}}
\label{eq:loss_total}
\end{align}
\end{subequations}
where \(D_{\text{pred}}\) and \(D_{\text{gt}}\) are the predicted and ground-truth depth maps, 
\(\mathcal{C}\) the contact region defined by the binary mask \(C(\mathbf{x})=1\), and \(\bar{\mathcal{C}}\) its complement. Contact masks are used only during training and are not required at inference.

\subsection{3D FORCE RECONSTRUCTION}
\label{sec:force_reconstruction}
We map 3D marker displacements to forces assuming linear elasticity. For small deformations, surface forces and displacements are related through the stiffness matrix as $\mathbf{F} = \mathbf{K}\,\mathbf{U}$. Each marker’s 3D position is recovered using its tracked image location, predicted normal displacement and camera matrix, and subtracted from its rest position to form the displacement vector $\mathbf{u}_j \in \mathbb{R}^3$. Stacking all displacement vectors yields the global deformation vector $\mathbf{U} \in \mathbb{R}^{3M}$. The gel is modeled as a linear, isotropic, elastic block discretized with cubic hexahedral elements and surface nodes aligned with markers, establishing a one-to-one correspondence between markers and top-layer FEM nodes. The bottom surface is fixed to enforce boundary conditions and unobserved nodes are eliminated using static condensation~\cite{bathe2006finite}, assuming zero external loads. The resulting matrix $\mathbf{K}\in\mathbb{R}^{3M\times3M}$ encodes the elastic couplings among the observed surface nodes. It is computed offline from the gel geometry and material parameters~\cite{ma2019dense}, and reused during online operation for fast inference.

\subsection{DATASET COLLECTION}
\label{sec:data_collection}

To train the normal deformation network and evaluate force estimation, we collect a dataset of controlled indentations using a CNC milling machine with a six-axis force–torque sensor (Bota Systems Rokubi) and our tactile sensor, all synchronized on a single computer. Each recording consists of a probing sequence, where the end-effector contacts the gel and performs the following motion pattern, Normal Down → Wait → Shear → Wait → Shear Opposite → Wait → Normal Up, with events, end-effector position, and forces logged continuously. We vary contact conditions across locations, speeds, shear directions, and four indenters. The dataset comprises 219 recordings (154/52/13 train/val/test), resulting in \(\approx 17\,\mathrm{min}\) of event data, and spanning normal and shear forces up to 20\,N and 4\,N.

Normal deformation is simulated from indenter geometry, indentation depth, and gel parameters by modeling the gel as a linear elastic isotropic, and homogeneous half-space. Under these assumptions, and purely normal contact between indenter and gel, normal deformation is approximated using the Boussinesq solutions \cite{boussinesq1885application}, offering a balance between computational efficiency and physical realism. The resulting deformation maps are projected to the camera frame and normalized to \([0,1]\), serving as ground truth for training.

\section{EXPERIMENTS}

All components operate at a fixed event rate, processing sequential windows of 20{,}000 events from each test set recording, ensuring that each input contains meaningful spatio-temporal information and avoiding cases with few or no events (e.g. during rest phases).

\begin{table}[t]
\centering
\caption{Lost tracks per patch size (mean $\pm$ std). Patch size is reported as multiple of the mean marker diameter.}
\label{tab:tracking_results}

\setlength{\tabcolsep}{6pt}    
\renewcommand{\arraystretch}{0.9}

\begin{tabular}{lccccc}
\toprule
Patch size & 1.0× & 2.0× & 3.0× & 4.0× & 5.0× \\
\midrule
Refinement     & 4.4±4.5 & 3.1±5.6 & \textbf{2.8±5.1} & 3.0±2.8 & 5.0±6.4 \\
No refinement  & 8.7±4.7 & 3.8±4.3 & 3.2±4.3 & 3.9±3.7 & 5.2±5.7 \\
\bottomrule
\end{tabular}
\vspace{-1.5em}
\end{table}

\subsection{MARKER TRACKING EVALUATION}
\label{sec:tracking_eval}

We evaluate the tracker following~\cite{funk2024evetac} using start–end consistency as proxy for error, since sequences begin and end in the undeformed state. A marker is considered lost if its displacement exceeds 15 pixels ($\approx$ marker diameter). We report mean lost tracks per sequence. As shown in Table~\ref{tab:tracking_results}, patch size strongly affects performance: small patches tend to lose markers, while large ones increase the likelihood of capturing events not generated by the corresponding marker, such as those caused by object edges or neighboring markers under large deformation cases. Intermediate sizes ($\approx$ 3.0$\times$ marker diameter) provide the best trade-off. Grid-aware refinement further improves robustness by enforcing local coherence; we use $\lambda=0.2$ as larger values can lead to over-smoothing of trajectories. Overall, the tracker achieves errors below one marker diameter ($\approx 0.5$\,mm). While limited to start–end consistency, this metric is a practical proxy given the lack of ground-truth trajectories. Obtaining ground-truth tracks is challenging with an event-only setup, though a hybrid RGB-event camera could enable deeper future evaluation.

\subsection{DEPTH ESTIMATION EVALUATION}
\label{sec:depth_eval}

\textbf{Effect of contact-aware loss.}
The proposed loss separates contact and non-contact regions to balance supervision and and make the network invariant to the indenters contact geometry. To evaluate its impact, we fix the accumulation time to 10~ms and the contact weighting parameter $\lambda_c=0.5$, and compare against a uniform $\mathcal{L}_1$ loss under identical settings. As shown in Table~\ref{tab:depth_ablation}, the contact-aware formulation improves accuracy in contact regions and reduces overall error. This could be particularly relevant for datasets with many small-contact cases and diverse indenter geometries.

\textbf{Effect of event accumulation time}.
The event accumulation window is a key parameter in event-based representations, as it determines the encoded motion information. Short windows capture fine dynamics but can be sparse under slow motion, while long windows provide richer history at the cost of information overlap, especially in highly textured scenes. We evaluate accumulation times from 5~ms to 70~ms (Table~\ref{tab:depth_ablation}) using the contact-aware loss with $\lambda_c=0.5$. Results show that overly long windows degrade accuracy and adopt 10~ms, which achieves the best performance.

\begin{table}[t]
\centering
\caption{Depth error (mm, mean $\pm$ std) for different accumulation times and loss functions.}
\label{tab:depth_ablation}

\setlength{\tabcolsep}{4pt}
\renewcommand{\arraystretch}{0.9}

\begin{tabular}{llccc}
\toprule
Loss & Time (ms) & Total & Contact & Non-contact \\
\midrule
\multirow{5}{*}{Contact-aware} 
 & 5  & 0.096$\pm$0.086 & 0.276$\pm$0.260 & 0.066$\pm$0.054 \\
 & 10 & \textbf{0.091$\pm$0.081} & \textbf{0.265$\pm$0.228} & 0.063$\pm$0.051 \\
 & 30 & 0.115$\pm$0.110 & 0.308$\pm$0.265 & 0.080$\pm$0.063 \\
 & 50 & 0.142$\pm$0.121 & 0.375$\pm$0.312 & 0.100$\pm$0.071 \\
 & 70 & 0.129$\pm$0.129 & 0.355$\pm$0.301 & 0.084$\pm$0.070 \\
\addlinespace[0.3em]
\cmidrule(lr){1-5}
Uniform 
 & 10 & 0.093$\pm$0.095 & 0.297$\pm$0.275 & \textbf{0.056$\pm$0.053} \\
\bottomrule
\end{tabular}
\vspace{-1.0em}
\end{table}

\textbf{Depth error per indentation depth range}.
We group test samples by ground-truth depth and compute per-group error using the best model. The model shows consistent performance across depths, with no significant depth-dependent bias or degradation at larger deformations, maintaining mean errors below 0.3~mm.

\vspace{-0.1em}
\subsection{FORCE ESTIMATION EVALUATION}
\label{sec:force_eval}

\begin{figure}[]
    \centering
    \includegraphics[width=\linewidth]{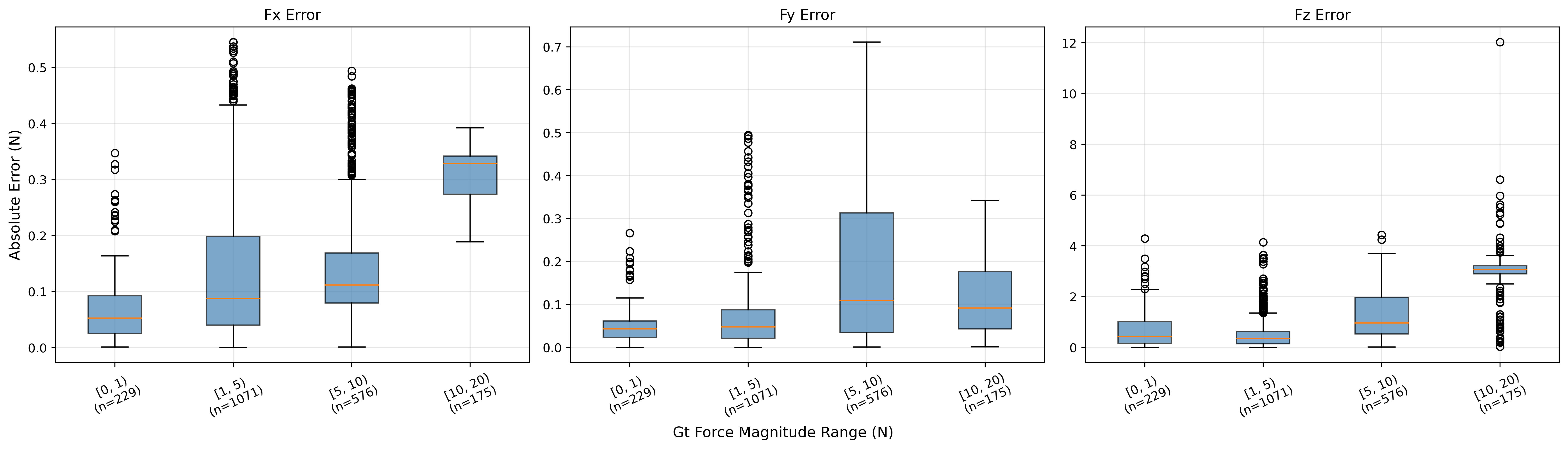}
    \caption[Force errors for different ground-truth force ranges.]{
   Force error distributions. Best performing model.}
    \label{fig:total_force_errors}
    \vspace{-1.5em}
\end{figure}

Due to the lack of dense force measurements, we follow~\cite{ma2019dense, zhang2022deltact} and evaluate by summing predicted per-marker forces and comparing them to measured net forces. We also assess spatial accuracy by comparing the true indenter position with the predicted contact location, computed as the center of mass of the estimated depth map. Error distributions across force ranges are shown in Fig.~\ref{fig:total_force_errors}, with results summarized in Table~\ref{tab:force_errors}. The method performs well across ranges. For shear forces, mean errors are below 0.4~N, while for normal forces they remain below 1~N except at the highest range. Unlike depth estimation, performance exhibits a range-dependent bias, likely due to deviations from linear elasticity at larger deformations. For force localization, the method achieves a mean error of 2.47 pixels ($<0.1$~mm in physical space), due to accurate depth estimation.

\begin{table}[t]
\centering
\caption{Force and contact position errors (mean $\pm$ std).}
\label{tab:force_errors}

\setlength{\tabcolsep}{4pt}
\renewcommand{\arraystretch}{0.9}

\begin{tabular}{lcc}
\toprule
 Axis & Force (N) & Position (px) \\
\midrule
$x$ & 0.14 $\pm$ 0.12 & 1.59 $\pm$ 4.62 \\
$y$ & 0.10 $\pm$ 0.13 & 1.55 $\pm$ 4.24 \\
$z$ & 0.93 $\pm$ 1.00 & -- \\
Total & 0.94 $\pm$ 1.00 & 2.47 $\pm$ 6.19 \\
\bottomrule
\end{tabular}
\vspace{-1.6em}
\end{table}

\subsection{RUNTIME PERFORMANCE}
\label{sec:runtime}
We run the pipeline at fixed event update rates of 10 ms, as constant-event processing is not best suited for real-time use. Tracking (C++) and force estimation (Python) run in parallel and only SAE construction depends on the event rate, which is not the bottleneck. With runtimes of 4.6~ms and 9.6~ms on average, the system achieves $\sim$100~Hz on a Xeon W-2235 CPU (3.80~GHz) and an NVIDIA RTX~3080 GPU.

\section{DISCUSSION \& CONCLUSION}
Our results show that dense 3D force fields can be reconstructed from sparse event measurements at high temporal resolution. However, some limitations remain. The linear elasticity assumption can introduce errors under large deformations, as observed in our experiments, and limited indenter diversity may restrict generalization to more complex contacts. Future work should explore non-linear force models, expand dataset coverage, and integrate event-based tactile sensing into downstream robotic tasks to better assess the benefits of high-frequency dense force feedback.

\addtolength{\textheight}{-12cm}   % This command serves to balance the column lengths

%%%%%%%%%%%%%%%%%%%%%%%%%%%%%%%%%%%%%%%%%%%%%%%%%%%%%%%%%%%%%%%%%%%%%%%%%%%%%%%%

%%%%%%%%%%%%%%%%%%%%%%%%%%%%%%%%%%%%%%%%%%%%%%%%%%%%%%%%%%%%%%%%%%%%%%%%%%%%%%%%

%%%%%%%%%%%%%%%%%%%%%%%%%%%%%%%%%%%%%%%%%%%%%%%%%%%%%%%%%%%%%%%%%%%%%%%%%%%%%%%%

\section*{ACKNOWLEDGMENTS}
We would like to thank Kirk Scheper for contributing to the tactile sensor prototype and Arjun Bhardwaj for his valuable feedback in the project.

%%%%%%%%%%%%%%%%%%%%%%%%%%%%%%%%%%%%%%%%%%%%%%%%%%%%%%%%%%%%%%%%%%%%%%%%%%%%%%%%

\vspace{-0.1em}
\bibliographystyle{IEEEtran}
\bibliography{references}

\end{document}